\documentclass[runningheads]{llncs}
\usepackage[T1]{fontenc}
\usepackage{graphicx}
\usepackage{amsmath,amsfonts}
\usepackage{algorithmic}
\usepackage{algorithm}
\usepackage{array}
\usepackage{textcomp}
\usepackage{stfloats}
\usepackage{url}
\usepackage{verbatim}
\usepackage{graphicx}
\usepackage{cite}
\usepackage{multirow}
\usepackage{bbding}
\usepackage[inkscapelatex=false]{svg}
\usepackage{subcaption}
\usepackage{booktabs}
\usepackage[misc]{ifsym}

\newcommand{\G}{{\mathcal{G}}}

\newcommand{\LL}{{\mathcal{L}}}

\begin{document}
\title{Semi-supervised Anomaly Detection with Extremely Limited Labels in Dynamic Graphs}


\author{Jiazhen Chen\inst{1} \and
Sichao Fu\inst{2}\Letter \and
Zheng Ma\inst{3}\and
Mingbin Feng\inst{1}\Letter \and
Tony S. Wirjanto\inst{1} \and
Qinmu Peng\inst{2}}
\institute{Department of Statistic and Actuarial Science, University of Waterloo \\
\email{\{j385chen, ben.feng, twirjanto\}@uwaterloo.ca} \and
School of Electronic Information and Communications, Huazhong University of Science and Technology \\ 
\email{fusichao$\_$upc@163.com, pengqinmu@hust.edu.cn} \and
Cheriton School of Computer Science, University of Waterloo \\
\email{z43ma@uwaterloo.ca}}

\authorrunning{Chen et al.}

\maketitle              
\begin{abstract}
Semi-supervised graph anomaly detection (GAD) has recently received increasing attention, which aims to distinguish anomalous patterns from graphs under the guidance of a moderate amount of labeled data and a large volume of unlabeled data. Although these proposed semi-supervised GAD methods have achieved great success, their superior performance will be seriously degraded when the provided labels are extremely limited due to some unpredictable factors. Besides, the existing methods primarily focus on anomaly detection in static graphs, and little effort was paid to consider the continuous evolution characteristic of graphs over time (dynamic graphs). To address these challenges, we propose a novel GAD framework (EL$^{2}$-DGAD) to tackle anomaly detection problem in dynamic graphs with extremely limited labels. Specifically, a transformer-based graph encoder model is designed to more effectively preserve evolving graph structures beyond the local neighborhood. Then, we incorporate an ego-context hypersphere classification loss to classify temporal interactions according to their structure and temporal neighborhoods while ensuring the normal samples are mapped compactly against anomalous data. Finally, the above loss is further augmented with an ego-context contrasting module which utilizes unlabeled data to enhance model generalization. Extensive experiments on four datasets and three label rates demonstrate the effectiveness of the proposed method in comparison to the existing GAD methods.

\keywords{Graph anomaly detection \and Semi-supervised learning \and Extremely limited labels \and Graph neural network \and Graph contrastive learning.}
\end{abstract}

\section{Introduction}
Anomaly detection (AD) is the process of identifying irregular patterns in data that deviate from expected behavior~\cite{chen2024towards}. In today's interconnected world, many datasets from various domains inherently form complex network structures~\cite{wei2021branch,yang2020tensor,yang2019survey,chen2022antibenford,tang2008arnetminer}. These network structures are often contaminated with anomalies. For instance, financial networks may exhibit irregular transactional patterns indicative of fraud, while social networks could contain spurious accounts engaged in malicious activities. In light of the inherent complexity of these graph-structured datasets and the severe consequences of ignoring anomalies, there has been a growing research emphasis on graph anomaly detection (GAD) in recent years.

Owing to the rarity of anomaly events in real-world applications, existing GAD methods primarily rely on unsupervised learning-based approaches. These methods typically utilize reconstruction-based~\cite{ding2019deep,luo2022comga,zhang2022unsupervised} and contrastive learning-based techniques~\cite{liu2021anomaly,zheng2021generative,zhang2022reconstruction,pan2023prem} to learn normal patterns within unlabeled data. Anomalies are then detected as deviations from these patterns. For example, reconstruction-based methods like DOMINANT~\cite{ding2019deep} and ComGA~\cite{luo2022comga} employ autoencoders to reconstruct graph structures and attributes, whereas anomalies are identified through substantial reconstruction errors. On the other hand, contrastive learning-based methods, such as COLA~\cite{liu2021anomaly} and PREM~\cite{pan2023prem}, detect anomalies by comparing nodes with their surrounding structures and identifying anomalies that have large inconsistencies highlighted via contrastive losses. However, these methods primarily focus on static graphs, failing to consider the dynamic nature of real-world graphs that are characterized by constantly evolving nodes, edges, and attributes. Such dynamic information can be crucial for identifying anomalies. For example, a sudden spike in account creation during off-peak hours could indicate inauthentic behavior, while large financial transfers during low-transaction periods may signal irregular activities.

Recently, detecting anomalies in dynamic graphs has received increasing research interest, which requires models capable of capturing both structural dependencies and temporal patterns \cite{zheng2019addgraph, cai2021structural, taddy}. For example, AddGraph~\cite{zheng2019addgraph} processes dynamic graphs as sequences of snapshots, using GCNs~\cite{kipf2017semi} and GRUs~\cite{chung2014empirical} to model these dependencies within the dynamic graph data; it also applies selective negative sampling by generating negative edges based on node degrees and uses a margin-based loss to ensure normal edges receive lower anomaly scores. Similarly, TADDY \cite{taddy} operates on discrete graph snapshots, utilizing a transformer model to learn both static graph patterns and their temporal evolution. A binary classifier is trained to distinguish anomalies by comparing real edges with pseudo-anomalous edges. However, these aforementioned methods are primarily unsupervised learning-based and also rely heavily on the quality of the pseudo labels, which may risk identifying noisy samples as anomalies due to the lack of prior anomaly knowledge. 

In practice, it is often possible to acquire a limited set of labeled examples. This has spurred increased attention towards semi-supervised GAD, which combines labeled data with unlabeled data to improve detection performance~\cite{wang2019semi, kumagai2021semi, meng2021semi, ding2021few, tian2023sad}. For instance, SemiADC~\cite{meng2021semi} uses generative adversarial networks to learn the feature distribution of normal nodes and trains a classifier to distinguish these normal features from labeled anomalies. The anomaly scores are computed by combining the classifier output with a temporal consistency score, which is derived from evaluating node behavior across consecutive graph snapshots. Similarly, SAD~\cite{tian2023sad} enhances the node representation learning with a temporal graph encoder that aggregates features across continuous time steps. The model is trained under a supervised deviation network which enforces significant differences between normal and abnormal nodes. It also introduces a pseudo-label-based contrastive learning component to improve performance. Despite the promising results of these semi-supervised dynamic GAD methods, they often hinge on the availability of an adequate number of labeled anomalies. In reality, the rarity and difficulty of accurately identifying anomalies make this assumption impractical in real-world applications, thereby hampering the effectiveness of these methods.

To address these challenges, we propose EL$^{2}$-DGAD, a novel GAD approach designed to enhance the robustness of existing methods for dynamic graph anomaly detection, especially under the constraints of limited labeled data. Our framework focuses on two critical aspects: improving feature learning in dynamic graph structures and strengthening the robustness of the loss function.

First, we introduce a transformer-based dynamic graph encoder designed to effectively capture both temporal and structural properties of dynamic graphs. Unlike existing methods that primarily focus on local dependencies using GNN-based approaches~\cite{tgat,tian2023sad}, our encoder leverages both local and global attention mechanisms. The former extracts fine-grained patterns from a node's neighborhood, while the latter aggregates these patterns across the graph, thereby enhancing the detection of subtle and extensive anomalies across the graph. Furthermore, we integrate continuous-time embeddings into the transformer's attention mechanisms, allowing our model to seamlessly incorporate temporal dynamics and learn precise, time-sensitive representations. This contrasts with discrete snapshot-based methods~\cite{zheng2019addgraph,liu2021anomaly,cai2021structural}, which lose precision by grouping events into fixed intervals, making it harder to detect time-sensitive anomalies.

Furthermore, to effectively utilize the limited labels, we propose a novel ego-context hypersphere classification loss. It learns a context-aware hypersphere boundary by considering the ``ego'' (target instance) and its ``context'' (historically evolving graph dynamics). 
By learning a robust dynamic boundary for abundant normal data, rather than modeling the irregular and diverse patterns of anomalies, we reduce the reliance on a large number of labeled anomalies. Instances that deviate from this learned dynamic boundary can be effectively identified as anomalies, even in scenarios with minimal supervision.
Lastly, to further improve model generalization, we integrate an ego-context contrastive loss, which ensures the consistency of neighboring information with target instances by utilizing the vast amount of unlabeled data.


To summarize, our main contributions include:

\begin{itemize}

\item \textbf{Novel Problem Setting:} To the best of our knowledge, we are the first attempt to investigate and address anomaly detection problem in dynamic graphs with extremely scarce labeling.
\item \textbf{Improving Dynamic Graph Feature Learning:} We propose a transformer-based encoder tailored for dynamic graphs. The encoder aims at capturing local and global structural dependencies while integrating continuous-time dynamics, thus enhancing the model's ability to detect subtle and time-sensitive anomalies within evolving graph structures.
\item \textbf{Enhancing Anomaly Detection with Limited Labels:} We propose an ego-context hypersphere classification loss to construct a robust and context-aware boundary for normal patterns, alongside an ego-context contrastive loss to leverage unlabeled data for improved generalization. These two losses jointly enable effective anomaly detection in dynamic graphs under extremely limited labeling conditions.
\item \textbf{Empirical Validation:} Extensive experiments on four datasets and three extremely low label rates demonstrate the effectiveness of the proposed method in comparison to existing GAD methods.

\end{itemize}

\section{Methodology}
\subsection{Problem Formulation}



In this paper, we define a dynamic graph as $\mathcal{G} = (\mathcal{V}, \mathcal{E})$, where $\mathcal{V}$ represents the set of nodes and $\mathcal{E}$ consists of temporal edges. Each edge, denoted by $\delta^t = (v_i, v_j, t, x_{ij})$, captures an interaction from node $v_i$ to node $v_j$ at time $t$, with $x_{ij}$ serving as the edge feature. To model the temporal dynamics effectively, we introduce two key subgraphs: the ego-graph $\mathcal{G}^t$, which includes all events up to and including time $t$, and the context graph $\mathcal{G}^{t^{-}}$, which contains all events up to time $t$ but excludes the event occurring at $t$. Specifically, our method learns to capture the consistency between the ego-graph and its historical context graph. If an event at time $t$ significantly deviates from this learned consistency, it is flagged as potentially anomalous. Our proposed EL$^{2}$-DGAD framework aims to detect anomalous edges in this dynamic graph setting, where each edge $\delta^t$ is associated with a latent label $y^t$, indicating whether the edge is normal ($y^t = 0$) or abnormal ($y^t = 1$). In this study, we have access to only a limited number of labeled edges, denoted as $Y^L$, while the majority of edges remain unlabeled ($Y^U$), with $|Y^L| \ll |Y^U|$. The goal is to develop an algorithm that assigns an anomaly score $s^t$ to each edge, represented as $f(\delta^t) = s^t$, to effectively distinguish between normal and anomalous edges.

\subsection{Overview}

\begin{figure*}[t]
    \centering
    \includegraphics[width=\linewidth]{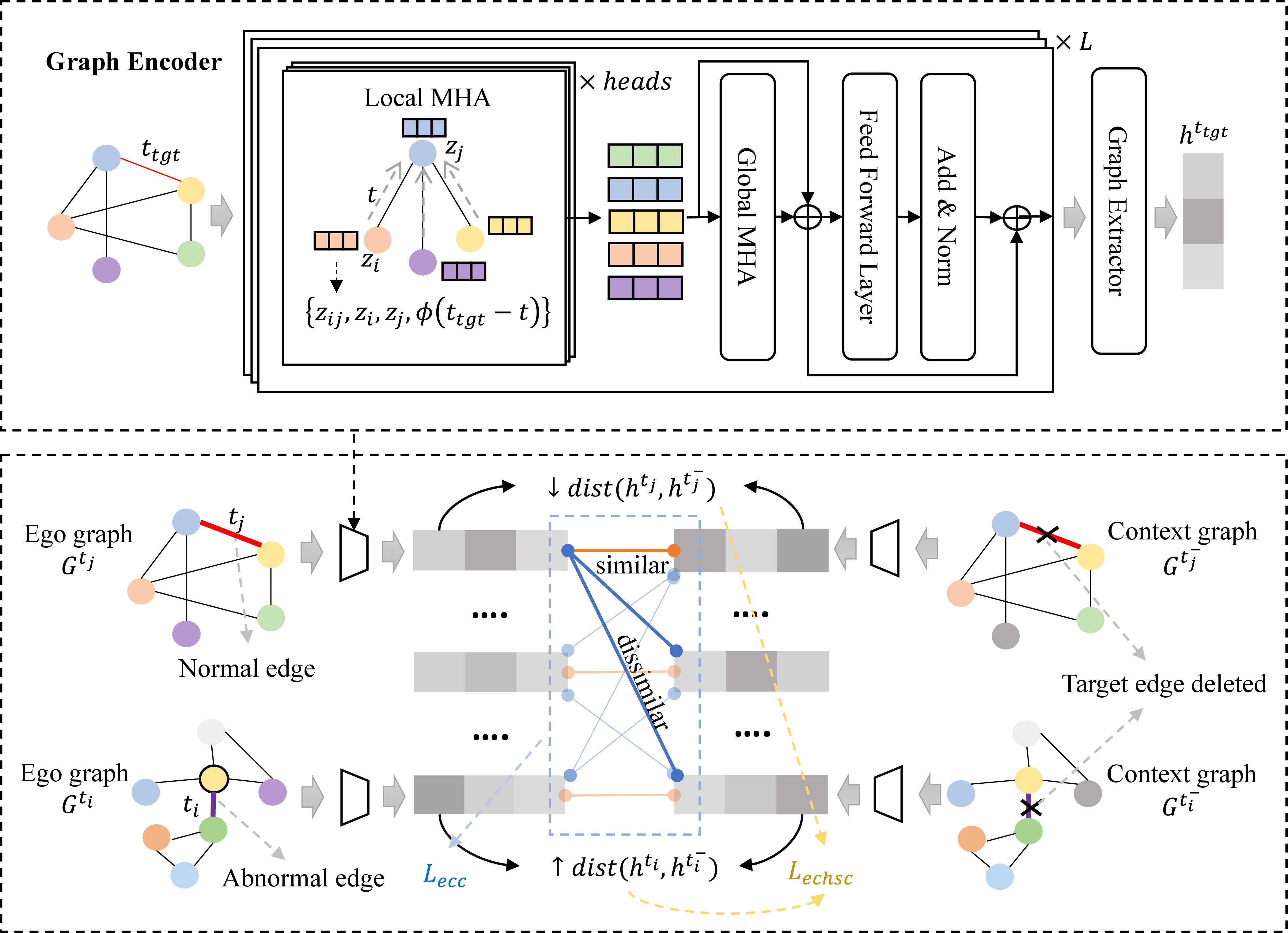}
    \caption{The architecture of EL$^{2}$-DGAD. The bottom section shows the ego and context graphs for two example edges (one normal and one abnormal), processed by the graph encoder in the top section. For a given edge, the distance between encoded ego and context graphs is minimized if the edge is normal and maximized if it is abnormal (via $\mathcal{L}^{ecc}$). Additionally, $\mathcal{L}^{echsc}$ regularizes all edges, enforcing consistency between each edge’s representation and its context graph. For the graph encoder, each layer includes a local MHA module that aggregates information from neighboring edges and nodes, along with the time difference relative to the edge’s timestamp. The local MHA outputs are then fed into a global MHA, followed by feed-forward networks and layer normalization.}
    \label{fig: EL2DGAD_framework}
\end{figure*}

The overview of EL$^{2}$-DGAD is shown in Figure~\ref{fig: EL2DGAD_framework}. Specifically, for each input sample, an ego and a context graph are simultaneously defined. Each of the graphs is encoded through a transformer-based dynamic graph encoder, which builts on top of a series of local and global attentions (Section~\ref{sec: graph_encoder}). During model training, we introduce an ego-context hypersphere classification loss in Section~\ref{sec: semi-supervised loss} that classifies edges by quantifying the similarity between their ego representation and the corresponding dynamic neighboring context. Finally, an ego-context contrastive loss in Section~\ref{sec: contrastive loss} is incorporated that is designed to mine and learn from the patterns of normality prevalent within the graph based on unlabeled data.

\subsection{Graph Encoder}
\label{sec: graph_encoder}
In scenarios with extremely limited labeled data, it is crucial to design a robust model that effectively captures the temporal and structural dependencies within dynamic graphs. These graphs are constantly evolving, and anomalies may manifest through unexpected temporal patterns or deviations from typical dynamic behaviors. For example, a node that is usually inactive suddenly exhibiting a surge in activity could indicate anomalous behavior. In addition to temporal patterns, both the influence of a node’s neighbors and the overall graph structure can reveal anomalies. Neighbor relationships capture local patterns within a node’s immediate vicinity, where anomalies might appear as atypical connections or attribute deviations. Conversely, graph-wide structures represent the global organization of the graph, where anomalies may disrupt community structures or introduce unexpected cross-community interactions. Therefore, a robust backbone model must integrate both local and global information while simultaneously capturing the temporal evolution of the graph.

To address these challenges, we propose a transformer-based dynamic graph encoder designed to effectively capture the temporal and structural properties of dynamic graphs. Drawing inspiration from recent advancements in graph transformers~\cite{ying2021transformers,chen2022structure}, our approach leverages local attention mechanisms to extract fine-grained patterns from a node’s neighborhood and global attention mechanisms to aggregate these patterns across the entire graph, thereby modeling both local and global dependencies. Furthermore, continuous-time embeddings are integrated into the attention mechanism, allowing the model to seamlessly incorporate temporal dynamics. 



Concretely, our graph encoder takes a subgraph at time $t$ as the input (e.g., $\G^{t}$ or $\G^{t^{-}}$) and generates node representations that reflect the dynamic status of the nodes. We start by aggregating 1-hop neighborhood information through masked multi-head self-attention, which we refer to as local MHA. To capture the evolving nature of dynamic graphs, we define the entries of $Q$, $K$, and $V$ at layer $l$ as follows:
\begin{align}
Q^l_{ij} = \text{MLP}(\text{Concat}(x_{ij}, z^{l-1}_i, z^{l-1}_j, \phi(\Delta_t)))W^{Q,l} \\
K^l_{ij} = \text{MLP}(\text{Concat}(x_{ij}, z^{l-1}_i, z^{l-1}_j, \phi(\Delta_t)))W^{K,l} \\
V^l_{ij} = \text{MLP}(\text{Concat}(x_{ij}, z^{l-1}_i, z^{l-1}_j, \phi(\Delta_t)))W^{V,l}
\end{align}
where $z^{l-1}_i$ is the output for node $i$ from layer $l-1$, and $z_i^0$ is a learnable embedding vectors specified by the degree of $v_i$, as detailed in~\cite{ying2021transformers}. MLP denotes the multi-layer perceptrons. $\Delta_t=t-t_{ij}$, where $t_{ij}$ is the occurring time of edge $(v_i, v_j)$. $\phi(\cdot)$ is the sinusoidal embedding function~\cite{transformer}, which generates a time representation by applying sine and cosine functions to the input time difference $\Delta_t$, scaled by a range of frequencies. 

Then, the attention output of node $i$ for a single head can be formulated as:
\begin{align}
    a^l_i= \sum_{j}\text{softmax}_j\left(\frac{Q^l_{ij}{K^l}^T_{ij}}{\sqrt{d}} M_{ij} \right)V^l_{ij}
\end{align}
where $M$ is the adjacency matrix of the input graph. Next, we fed the node representation obtained from the local MHA, as the input to the regular multi-head self-attention in~\cite{transformer}, which we referred to as the global MHA. Local MHA focuses on learning neighborhood information which is similar to GNNs but with their inherent limitations, whereas global attention broadens the perspective without explicit structural reliance. Thus, this combination enhances the expressiveness of our model in representing graph structures, effectively balancing the strengths of local and global information processing. 

Finally, the output of layer $l$ is obtained as
\begin{align}
    z^l=\text{LN}(\text{FFN}(z^{l-1, loc}+z^{l-1,glo}))+z^{l-1, loc}+z^{l-1, glo}
\end{align}
where $z^{l-1, loc}$ and $z^{l-1, glo}$ are the output from the local MHA and global MHA, LN and FFN denote the layer normalization~\cite{lei2016layer} and the feed-forward blocks respectively.




\subsection{Model Training}

\subsubsection{Ego-context HSC loss} 
\label{sec: semi-supervised loss}

To train the model, we incorporate the hypersphere classification (HSC) loss~\cite{ruff2020rethinking}, which is well-suited for anomaly detection tasks where labeled anomaly data is limited. Traditional classification approaches often assume that similar data points cluster naturally, but this assumption does not hold for anomalies, which typically do not form distinct clusters~\cite{chandola2009anomaly}. The HSC loss addresses this limitation by constructing a decision boundary in the form of a hypersphere, tailored for distinguishing between normal and anomalous data.


Under our context, the HSC loss of a given sample at $t$ is defined as:
\begin{align}
\LL_t^{hsc} = -y^t \log{l(x^t)} - (1-y^t) \log{(1-l(x^t))}
\label{eq. hsc}
\end{align}
where $l(z)=\exp{(-\|z\|^2)}$, $x^t$ and $y^t$ are the latent representation and label of an interaction event at time $t$. 
This formulation is a variant of the cross-entropy loss but with a unique focus: it constructs a spherical decision boundary that compacts normal samples towards a fixed center (the zero vector), while pushing anomalies away by maximizing their distance from this center. In essence, the goal is to pull normal data points closer to the origin, creating a compact representation, and to push out anomalous points to better distinguish them. The final classification output is given by $1-l(x^t)$, which represents the probability of an interaction being classified as normal.


To better align with the dynamic nature of graphs, where anomalous interactions are defined relative to their evolving environment, we modify the original HSC loss to explicitly incorporate temporal and structural context. Specifically, we represent the interaction at time $t$ using two graph perspectives: the ego graph $\G^t$, which captures the evolving interactions including the event at $t$, and the context graph $\G^{t-}$, which reflects the recent neighborhood structure up to but not including $t$. 

The latent representation for the interaction at time $t$ is then defined as the difference between these two graph perspectives:
\begin{align}
x^t = h^t - h^{t^{-}} 
\label{eq. distance of two graphs}
\end{align}
where $h^t$ and $h^{t^{-}}$ are the graph representations for ego and context graphs, respectively. Here, $h^t$ is computed as
\begin{align}
h^t = f_o(\text{concat}(z_i^t, z_j^t))
\label{eq. output_layer}
\end{align}
where $f_o$ is a two-layer MLP and $z_i^t$ is the outputs of node $i$ for $\G^t$ from the graph encoder. A separate graph encoder with different weights is used to compute $h^{t-}$ for the context graph.

By replacing the original latent representation in the HSC loss with $h^t-h^{t-}$, we formulate the ego-context HSC loss as
\begin{align}
\LL_t^{echsc} =& \ y^t\|h^t - h^{t^{-}}\|^2 - 
  (1-y^t)\log(1-\exp{(-\|h^t - h^{t^{-}}\|^2)})
    \label{eq. ego-context loss}
\end{align}

This modified loss directly measures how the addition of an interaction affects the deviation of the graph representation from the scenario without the interaction. Essentially, $\LL_t^{echsc}$ classifies interactions based on how well an edge aligns with the established patterns in the evolving graph structure from the past. 



\subsubsection{Ego-context Contrastive Loss} 
\label{sec: contrastive loss}

To mitigate the risk of overfitting to a limited number of labeled samples, we enhance the proposed framework with a contrastive learning component by utilizing the abundant unlabeled data. Here, we aim to harness the rich information in the unlabeled samples, thereby enhancing the ability of the model to discern between normal and anomalous patterns. A straightforward adaptation of the current framework is to maximize the similarity within the ego-context graph pairs from the same edge while minimizing the similarity for pairs from different edges. The underlying assumption is that the majority of edges are aligned with their context, given that the anomalies are rare in real-world situations. Specifically, the loss for an edge occurring at time $t$ is defined as:
\begin{align}
\LL^{ecc}_t= -\log(S(f_p(h^t), f_p(h^{t^{-}}))) - \log(1-S(f_p(h^t), f_p(h^{t^{-}_{neg}})))
\end{align}
where $h^{t^{-}_{neg}}$ represents the context graph representation of a different edge at $t_{neg}$ that is selected randomly within a training batch. The function $S(\cdot, \cdot)$ computes the cosine similarity between two inputs and applies a linear mapping to convert the value within 0 and 1. We use the cosine similarity instead of the one based on Euclidean distance as in Eq.~\ref{eq. ego-context loss} to let the model focus on learning the normal patterns from a different angle without compromising its ability to identify anomalies through the semi-supervised loss. The use of $f_p(\cdot)$, an MLP projector function, aligns with standard practices in contrastive learning which ensures that the unsupervised learning framework does not overshadow the primary task. 

Finally, the overall training loss is defined as:
\begin{align}
    \LL = \LL^{echsc} + \lambda \LL^{ecc}
\end{align}
where $\lambda$ is a hyper-parameter controlling the contribution of the contrastive loss.

\section{Experiments}


\begin{table}[t]
    \centering
    \caption{Summary statistics for datasets.} 
    \begin{tabular}{c|c|c|c} \hline
               Dataset & \# Nodes & \# Edges & \# Anomalies \\ \hline
               UCI  &  1899         &  14253   & 415      \\
               Digg & 29955 & 87709 & 2554 \\
               Reddit & 10984 & 672447 & 366 \\
               Wikipedia & 9227 & 157474 & 217 \\
   \hline
    \end{tabular}
    \label{tab: data_statistics_el2-dgad}
\end{table}

\subsection{Datasets}


We evaluate our approach on four dynamic network datasets: UCI~\cite{opsahl2009clustering}, Digg~\cite{de2009social}, Wikipedia~\cite{kumar2019predicting}, and Reddit~\cite{kumar2019predicting}. The Wikipedia dataset tracks user interactions, where each interaction corresponds to a user editing a wiki page, with the network evolving as users contribute over time. Dynamic labels indicate whether a user is banned from posting. The UCI Messages dataset represents a social network from an online community of students at the University of California, Irvine, where each node denotes a user, and edges signify exchanged messages. The Digg dataset is collected from the social news website Digg.com, which captures user interactions with nodes representing users and edges indicating replies between them. The Reddit dataset is a dynamic network that tracks active users posting across various subreddits, with each interaction representing a user posting in a specific subreddit. Dynamic labels indicate whether a user is banned from posting in a particular subreddit.

Since UCI and Digg do not contain labeled anomalous edges, we introduce synthetic anomalies using an anomaly injection procedure, following a similar setting as in~\cite{taddy}. The process involves clustering nodes into multiple clusters using spectral clustering based on the graph's structure and then generating a small portion of anomalous edges by randomly connecting nodes from different clusters, ensuring these connections are absent in the original graph. $3\%$ of anomalies are injected to reflect the realistic scenario where anomalies are typically rare. To seamlessly integrate these synthetic anomalies into the dynamic setting, the generated edges are uniformly inserted at random positions among the original edges. Each anomalous edge is assigned a timestamp by randomly selecting a value between the nearest preceding and succeeding normal edges' timestamps, thus maintaining temporal consistency with the original data.

The datasets are chronologically divided into training, validation, and testing sets based on interaction timings, with a split ratio of 50/20/30. This approach ensures that the model is trained on earlier interactions, validated on more recent data, and tested on the latest interactions, which mirrors real-world scenarios where future data is unknown at training time. In our experiments, we focus on evaluating performance under extremely limited label conditions by providing only 1, 2, or 3 labeled anomaly examples. For the labeled normal samples, we retain a proportion that matches the ratio of labeled anomalies to the total number of anomalies. The summary statistics of all datasets is shown in Table~\ref{tab: data_statistics_el2-dgad}.
\begin{table*}[t]
\caption{Experiment comparison with the state-of-the-art GAD methods in terms of AUC. The best results are marked in blue.}
  \centering
  \resizebox{\linewidth}{!}{
  \begin{tabular}{c|ccc|ccc|ccc|ccc}
    \hline
     Dataset & \multicolumn{3}{c|}{Wikipedia} & \multicolumn{3}{c|}{UCI} & \multicolumn{3}{c|}{Digg} & \multicolumn{3}{c}{Reddit} \\
    \cline{1-13}
     Labels & 1 & 2 & 3 & 1 & 2 & 3 & 1 & 2 & 3 & 1 & 2 & 3 \\
    \hline

    AddGraph & 0.631 & 0.641 & 0.663 & 0.732 & 0.732 & 0.754 & 0.739  & 0.740 & 0.742 & 0.578 & 0.580 & 0.582\\TGAT & 0.660 & 0.674 & 0.694 & 0.764 & 0.790 & 0.808 & 0.800 & 0.804 & 0.812 & 0.527 & 0.541 & 0.555 \\GDN & 0.516 & 0.529 & 0.533 & 0.510 & 0.546 & 0.547 & 0.511 & 0.510 & 0.513 & 0.532 & 0.531 & 0.533 \\SL-GAD & 0.467 & 0.524 & 0.526 & 0.503 & 0.497 & 0.502 & 0.494 & 0.497 & 0.504 & 0.503  &0.505  &0.500  \\TADDY & 0.616 & 0.617 & 0.618 & 0.769 & 0.765 & 0.772 & 0.811 & 0.841 & 0.842 & 0.516  &0.517  & 0.518 \\
    SAD & 0.661 & 0.674 & 0.710 & 0.774 & 0.798 & 0.816 & 0.809 & 0.820 & 0.841 & 0.561  & 0.566 & 0.592
    \\PREM & 0.549 & 0.555 & 0.555 & 0.531 & 0.532 & 0.536 & 0.512 & 0.513 & 0.515  & 0.516 & 0.535 & 0.536\\ \hline 
    EL$^2$-DGAD & \textcolor{blue}{\textbf{0.722}} & \textcolor{blue}{\textbf{0.725}} & \textcolor{blue}{\textbf{0.732}} & \textcolor{blue}{\textbf{0.839}} & \textcolor{blue}{\textbf{0.840}} & \textcolor{blue}{\textbf{0.842}} & \textcolor{blue}{\textbf{0.839}} & \textcolor{blue}{\textbf{0.844}} & \textcolor{blue}{\textbf{0.846}} & \textcolor{blue}{\textbf{0.611}} & \textcolor{blue}{\textbf{0.615}} & \textcolor{blue}{\textbf{0.616}} \\

  \hline
  \end{tabular}}
  \label{tab: comparison_results}
\end{table*}

\subsection{Model Implementation}

To construct the ego-graph $\mathcal{G}^t$, we include all edges with timestamps up to and including time $t$, capturing the interactions leading up to that specific time point. For the context graph $\mathcal{G}^{t^{-}}$, we similarly include all edges up to time $t$, but explicitly exclude any edge that occurs exactly at time $t$. This differentiation ensures that the context graph only represents historical information, without the current event. To manage the potentially large sizes of $\mathcal{G}^t$ and $\mathcal{G}^{t^{-}}$, we implement a sampling strategy that retains a subset of neighboring nodes for each edge endpoint. Specifically, we randomly sample 25, 10, and 5 neighboring nodes at the first, second, and third hops, respectively, with the condition that each sampled neighbor event must have a timestamp no later than the target edge’s timestamp, ensuring that only past interactions are considered. 

In addition, our model uses a two-layer transformer with 4 attention heads, and the hidden size is set to 128 across all modules. The contrastive loss weight $\lambda$ is set to 0.01 for Digg, UCI, Reddit and 10 for Wikipedia. The models are trained for 20 epochs using the Adam optimizer~\cite{kingma2014adam}, with a learning rate of 0.0001 for the Digg and UCI datasets, and 0.001 for the Wikipedia and Reddit dataset.



\subsection{Baselines and Performance Evaluation}


We benchmark our proposed method against several recent GAD models, namely GDN~\cite{ding2021few}, SL-GAD~\cite{zheng2021generative}, PREM~\cite{pan2023prem}, AddGraph~\cite{zheng2019addgraph}, TADDY~\cite{taddy}, TGAT~\cite{tgat}, and SAD~\cite{tian2023sad}. Among these, GDN, SL-GAD, and PREM are designed for static graph anomaly detection, while AddGraph, TADDY, TGAT, and SAD are tailored for dynamic graph settings. 



Note that only GDN, TADDY, AddGraph, and SAD can utilize label information by design. For a fair, label-inclusive comparison for other models, we perform a similar setting as in~\cite{tian2023sad}: applying a cross-entropy loss to the edge representations, which are derived from the concatenated node embeddings of the edge endpoints. Specifically, these edge representations are computed using a MLP network, as outlined in Eq.~\ref{eq. output_layer}. Thus, the performance of all models can be evaluated based on the classification results.  

\subsection{Comparison with State-of-the-art GAD}

Extensive results in Table~\ref{tab: comparison_results} show that our EL$^{2}$-DGAD consistently surpasses all comparison methods across four datasets, particularly in scenarios where labeled anomalies are extremely limited. The results highlight the effectiveness of our approach in incorporating both temporal and structural context, enabling the model to maintain robustness even under severe label scarcity. 

For models like GDN, PREM, and SL-GAD, which do not incorporate temporal information into their design, the decline in performance underscores the critical importance of capturing temporal dependencies in dynamic graph anomaly detection. Although AddGraph and TADDY are designed for dynamic graphs, their performance still lags behind ours by a significant margin. These methods rely on discrete graph snapshots, where timestamps of events within a single snapshot are assumed to be identical. Consequently, this can miss important details about the continuous evolution of the graph's structure. Moreover, they lack a robust semi-supervised learning mechanism, making it challenging to effectively utilize the limited labeled data. In contrast, our approach operates directly on continuous dynamic graphs, with the transformer-based encoder attending to the most relevant neighbors based on fine-grained temporal information. Additionally, we introduce an ego-context hypersphere classification loss that enforces the alignment of each edge with its immediate surrounding temporal and structural context. Together, these features enable EL$^2$-DGAD to more efficiently utilize the continuously evolving graph patterns, making it better suited to detect anomalies that are difficult to align with the normal behavior.

While SAD is also a semi-supervised anomaly detection framework designed for dynamic graphs, it exhibits a notable drop in performance under extreme label scarcity. Specifically, SAD relies on a deviation network as the main supervised learning component, using it to generate pseudo-labeled data that guides a separate contrastive learning process. However, when only a few labeled examples are available, the quality of these pseudo-labels declines, undermining the effectiveness of the contrastive learning component. In contrast, our approach decouples the unsupervised learning from the supervised training, allowing the model to retain its ability to generalize effectively even under extremely limited conditions. Furthermore, our approach offers additional advantages. First, the ego-context hypersphere classification loss provides a more precise instance-wise contextual comparison, compared to the deviation network used in SAD, which relies on comparisons to only global statistics. Second, our model employs a more robust transformer-based encoder, compared to the GAT encoder used in SAD, allowing it to more effectively capture temporal interactions between nodes from both local and global perspectives.

\subsection{Ablation Experiments}


Table~\ref{tab:ablation_results} presents the results of the ablation study. Overall, EL$^{2}$-DGAD achieves optimal performance when all components are included. Removing the local MHA leads to a significant performance drop, highlighting its essential role in capturing neighborhood structures. Similarly, omitting the global MHA also results in a noticeable decrease in performance, underscoring the importance of capturing global node interactions.
The exclusion of fine-grained timestamp integration within the Transformer’s attention mechanism also considerably degrades performance, especially in the Wikipedia dataset, where a node-to-node interaction can occur at many distinct timestamps. Ignoring these timestamps results in large information loss, as it prevents the model from capturing the temporal dynamics necessary to detect time-sensitive anomalies effectively. This further emphasizes the importance of modeling temporal changes in dynamic graphs. Additionally, the drop in performance without the contrastive loss component $\mathcal{L}^{ecc}$ indicates the value of leveraging unlabeled data to improve anomaly detection accuracy. Lastly, removing the ego-context contrast from $\mathcal{L}^{echsc}$ leads to a notable decline in performance. Without ego-context contrast, the model no longer minimizes the distance between each ego graph and its context graph; instead, all normal instances are pulled toward a single center, imposing a stronger regularization. This limits the model's ability to capture the diverse characteristics of normal instances, thereby reducing its ability to distinguish anomalies from normal events.

\begin{table}[t]
\caption{Performance evaluation on ablation experiments in terms of AUC. $w/o$ denotes without a certain module.}
  \centering
  \resizebox{1\linewidth}{!}{
  \begin{tabular}{c|ccc|ccc|ccc}
    \hline
    Dataset & \multicolumn{3}{c|}{Wikipedia} & \multicolumn{3}{c|}{UCI}  & \multicolumn{3}{c}{Digg} \\ \cline{1-10}
    
     Labels & 1 & 2 & 3 & 1 & 2 & 3 & 1 & 2 & 3  \\
     
    \hline
    $w/o$ Local MHA & 0.563 & 0.569 & 0.582 & 0.771 & 0.775 & 0.772 & 0.808 & 0.796 & 0.802 \\
    $w/o$ Global MHA & 0.694 & 0.709 & 0.716 & 0.835 & 0.836 & 0.838 & 0.802 & 0.807 & 0.800 \\
    $w/o \ \phi(t)$ in MHA & 0.465 & 0.473 & 0.474 & 0.789 & 0.791 & 0.800 & 0.719 & 0.730 & 0.756 \\
    $w/o$ $\LL^{ecc}$ & 0.653 & 0.683 & 0.683 & 0.838 & 0.837 & 0.840 & 0.815 & 0.821 & 0.821 \\
    $w/o$ ego-context in $\LL^{echsc}$ & 0.650 & 0.683 & 0.704 & 0.824 & 0.824 & 0.837 & 0.730 & 0.745 & 0.745 \\ \hline
    EL$^{2}$-DGAD &  \textcolor{blue}{\textbf{0.722}} &  \textcolor{blue}{\textbf{0.725}} &  \textcolor{blue}{\textbf{0.732}} &  \textcolor{blue}{\textbf{0.839}} & \textcolor{blue}{\textbf{0.840}}& \textcolor{blue}{\textbf{0.842}}& \textcolor{blue}{\textbf{0.839}}& \textcolor{blue}{\textbf{0.844}}& \textcolor{blue}{\textbf{0.846}} \\ \hline
  \end{tabular}}
\label{tab:ablation_results}
\end{table}

\subsection{Parameters Sensitivity}

\begin{figure}[t]
\begin{subfigure}[b]{1\linewidth}
    \centering
    \includegraphics[height=0.7cm]{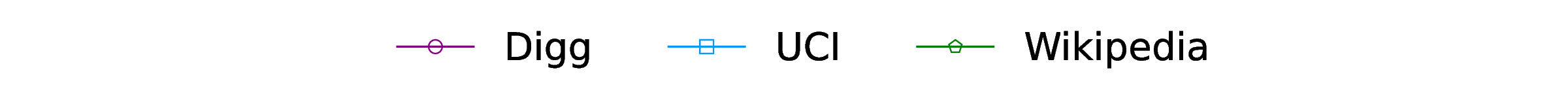}
\end{subfigure}
\newline
\begin{subfigure}[b]{0.49\linewidth}
    \centering
    \includegraphics[height=4.2cm]{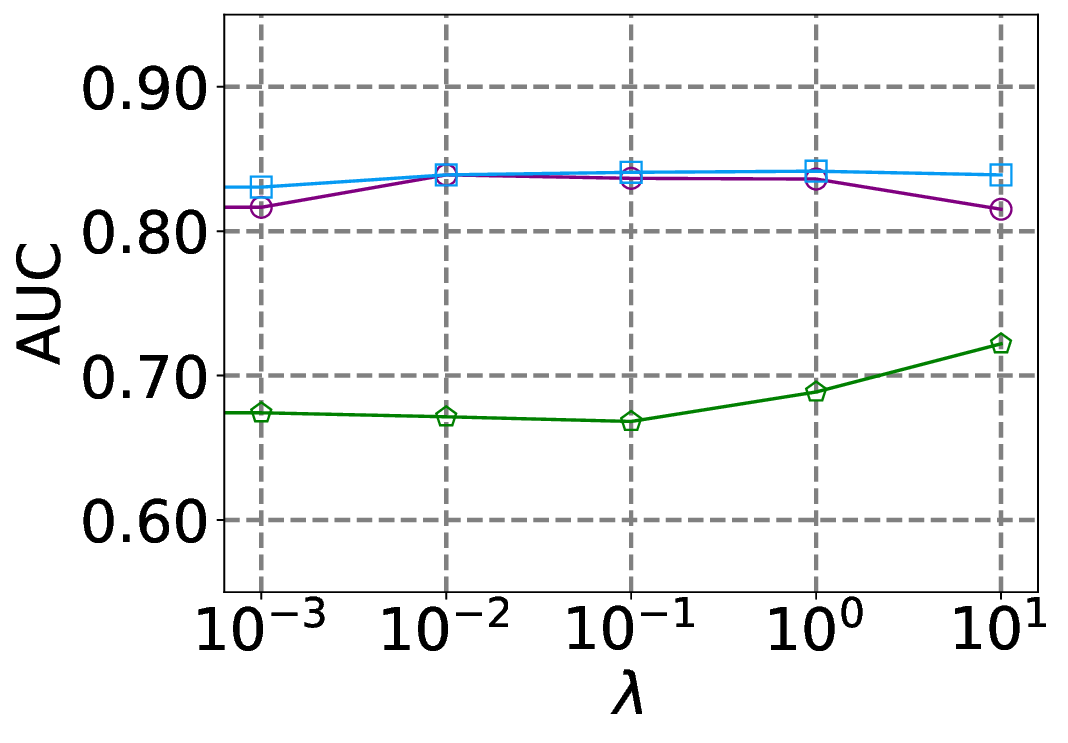}
    \caption{AUC vs. balance parameter $\lambda$}
     \label{fig: contra_results}
\end{subfigure}
\hfill
\begin{subfigure}[b]{0.49\linewidth}
    \centering
    \includegraphics[height=4.2cm]{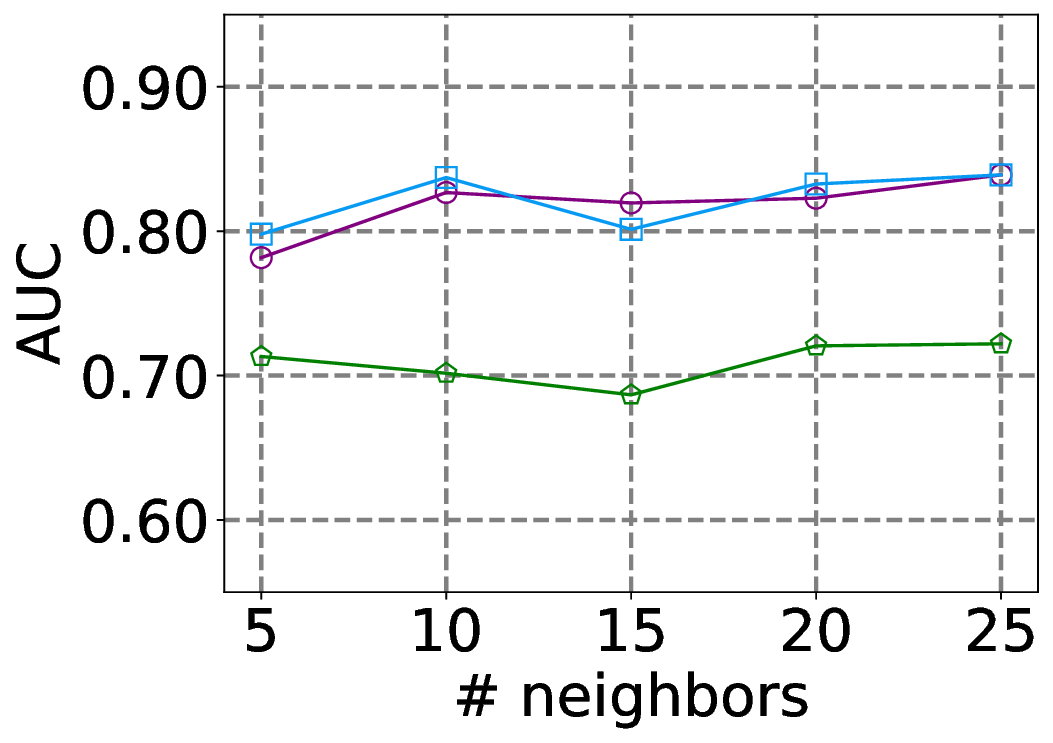} 
    \caption{AUC vs. number of neighbors}
    \label{fig: neighbor_results}
\end{subfigure}
\caption{Sensitivity analysis of $\lambda$ and neighbors number when there is one labeled anomaly available in the training dataset.}
\label{fig: sensitivity_results}
\end{figure}

Figure~\ref{fig: sensitivity_results} illustrates the AUC performance for three datasets (Digg, UCI, and Wikipedia) as it varies with changes in the balance parameter $\lambda$ for the ego-context contrastive learning component and the number of neighbors in the first hop. In Figure 2(a), the Wikipedia dataset shows improved AUC scores with higher $\lambda$ values compared to Digg and UCI. This may be due to the greater complexity of anomalies in Wikipedia, making it a more challenging task under extremely limited conditions. Consequently, increasing $\lambda$ enhances the model’s ability to capture general patterns of normal samples, which improves the robustness of the model. 

In Figure 2(b), AUC performance for all three datasets slightly increases as the number of neighbors rises, indicating that additional neighborhood information helps the model better distinguish between normal and anomalous interactions within the graphs. Overall, the model’s sensitivity to changes in $\lambda$ and the number of neighbors remains relatively low, suggesting that it is robust to variations in these hyperparameters.

\subsection{t-SNE Visualization} 

\begin{figure}[t]
\begin{subfigure}[b]{0.495\linewidth}
    \centering
    \includegraphics[height=3.9cm]{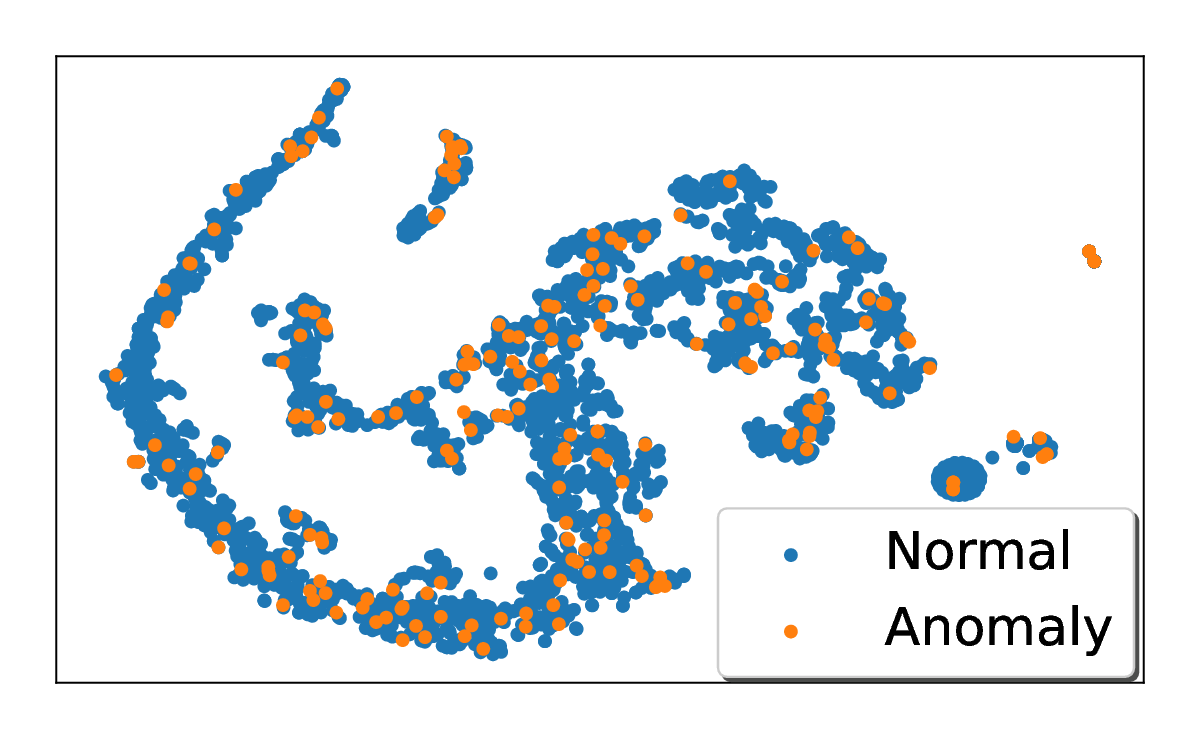}
    \caption{TGAT}
    \label{fig: tgat_tsne}
\end{subfigure}
\begin{subfigure}[b]{0.495\linewidth}
    \centering
    \includegraphics[height=3.9cm]{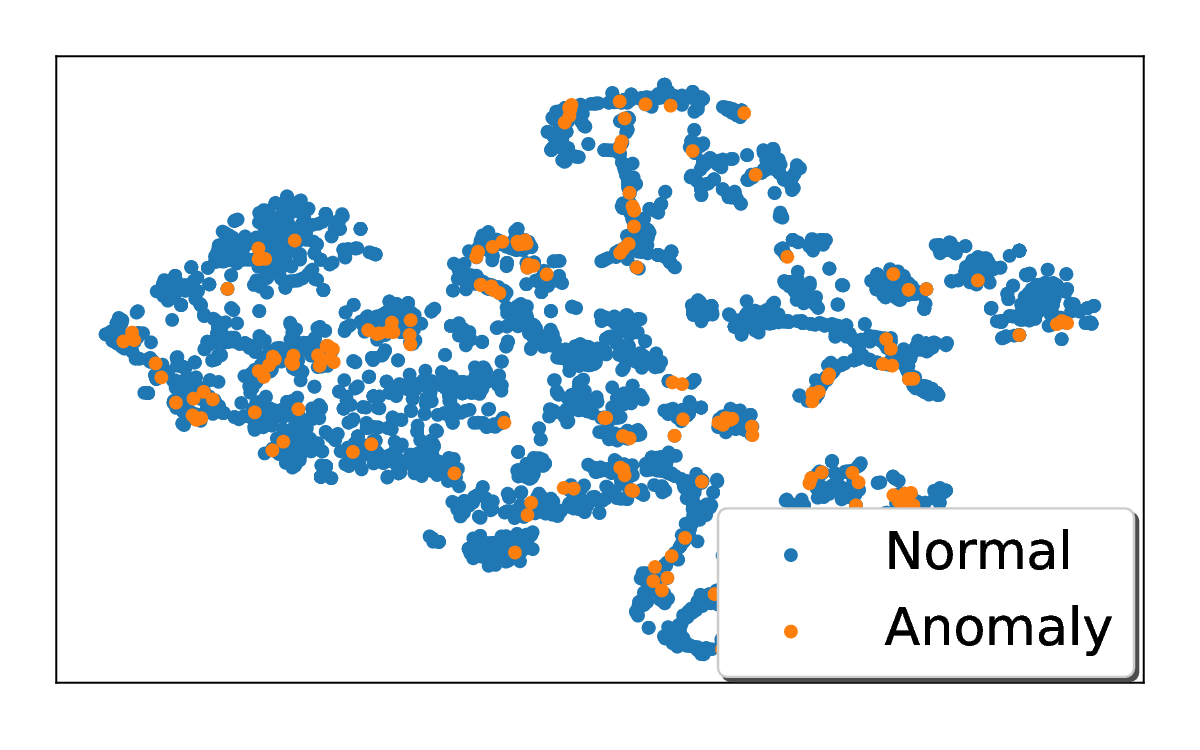}
    \caption{AddGraph}
     \label{fig: addgraph_tsne}
\end{subfigure}
\newline
\begin{subfigure}[b]{0.495\linewidth}
    \centering
    \includegraphics[height=3.9cm]{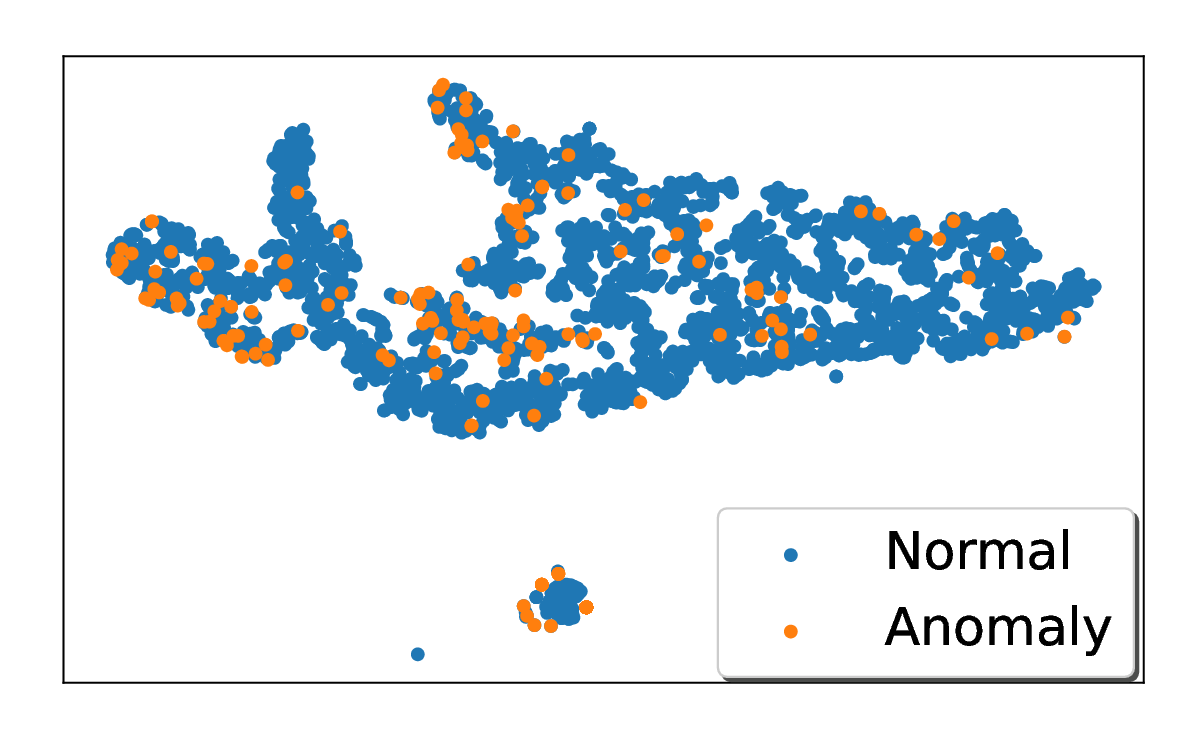}
    \caption{SAD}
    \label{fig: sad_tsne}
\end{subfigure}
\begin{subfigure}[b]{0.495\linewidth}
    \centering
    \includegraphics[height=3.9cm]{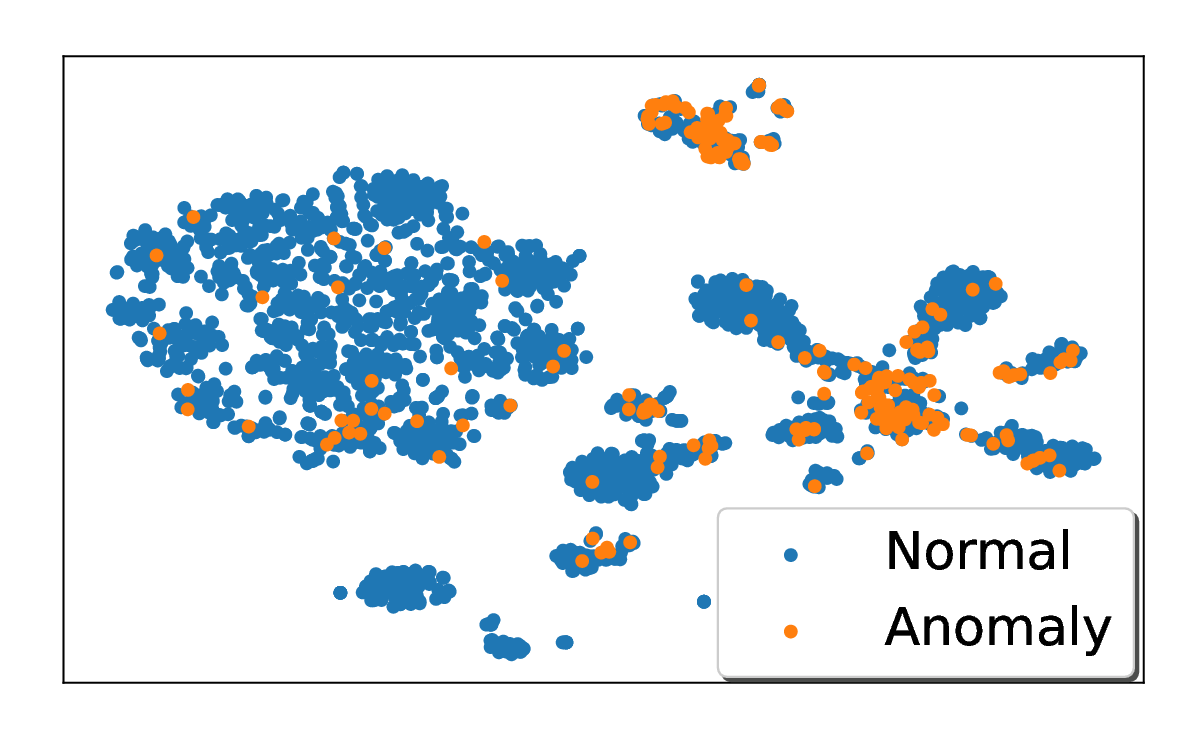}
    \caption{EL$^{2}$-DGAD (ours)}
     \label{fig: ours_tsne}
\end{subfigure}
\caption{t-SNE Visualization on the Wikipedia dataset.}
\label{fig: visualization}
\end{figure}

Figure~\ref{fig: visualization} shows t-SNE visualizations of the output embeddings from the layer before anomaly scoring in the Wikipedia dataset, where the selected four models were trained with supervision from only one labeled anomaly. In these visualizations, blue points represent normal instances, while orange points denote anomalies. In the embeddings produced by TGAT and AddGraph, anomalies are dispersed throughout the normal samples, indicating a limited capacity for distinguishing between normal and anomalous instances. Similarly, SAD shows slight improvement with a somewhat better separation, but anomalies remain scattered and intermingled within the normal clusters, lacking a clear boundary.

In contrast, our proposed EL$^{2}$-DGAD exhibits a relatively clearer separation, with anomalies forming distinct clusters that are well-separated from normal samples. This distinct clustering suggests that EL$^{2}$-DGAD more effectively learns discriminative embeddings, allowing it to capture the subtle distinctions between normal and anomalous interactions. This improved separation highlights the robustness and effectiveness of our approach for the GAD task, particularly in the challenging scenario of training with extremely limited labeled anomalies.

\section{Conclusion}

In this paper, we presented EL$^{2}$-DGAD, the first framework designed to address the anomaly detection problem in dynamic graphs under the challenging condition of extremely limited labeled data. Our approach integrates a transformer-based dynamic graph encoder that captures evolving graph patterns from both local and global perspectives, leveraging continuous-time embeddings to preserve temporal precision. This design is further complemented by an ego-context hypersphere classification loss and an unsupervised ego-context contrastive loss, which together effectively utilize the limited labeled anomalies while harnessing the rich information from abundant unlabeled data. Through extensive experiments on multiple benchmark datasets across varying label rates, EL$^{2}$-DGAD demonstrates significant performance improvements over existing graph anomaly detection methods. The results validate the effectiveness of our model in accurately detecting anomalies across diverse dynamic graph scenarios, even under severe label constraints.

\section*{Acknowledgements}

This work was supported in part by the National Key Research and Development Program of China under Grant 2022YFF0712300, in part by the Fundamental Research Funds for the Central Universities under Grant YCJJ20241203.

\bibliographystyle{splncs04} 
\bibliography{mybib} 

\end{document}